# Unveiling theory of mind in large language models: A parallel to single neurons in the human brain

Mohsen Jamali[1], Ziv M. Williams[1,2,3*], Jing Cai[1*†]

[1] Department of Neurosurgery, Massachusetts General Hospital, Harvard Medical School, Boston, MA.
[2] Harvard-MIT Division of Health Sciences and Technology, Boston, MA.
[3] Harvard Medical School, Program in Neuroscience, Boston, MA.

[*] Senior co-authors
[†] Correspondence should be sent to jcai5@mgh.harvard.edu

## Abstract

With their recent development, large language models (LLMs) have been found to exhibit a certain level of Theory of Mind (ToM), a complex cognitive capacity that is related to our conscious mind and that allows us to infer another's beliefs and perspective. While human ToM capabilities are believed to derive from the neural activity of a broadly interconnected brain network, including that of dorsal medial prefrontal cortex (dmPFC) neurons, the precise processes underlying LLM's capacity for ToM or their similarities with that of humans remains largely unknown. In this study, we drew inspiration from the dmPFC neurons subserving human ToM and employed a similar methodology to examine whether LLMs exhibit comparable characteristics. Surprisingly, our analysis revealed a striking resemblance between the two, as hidden embeddings (artificial neurons) within LLMs started to exhibit significant responsiveness to either true- or false-belief trials, suggesting their ability to represent another's perspective. These artificial embedding responses were closely correlated with the LLMs' performance during the ToM tasks, a property that was dependent on the size of the models. Further, the other's beliefs could be accurately decoded using the entire embeddings, indicating the presence of the embeddings' ToM capability at the population level. Together, our findings revealed an emergent property of LLMs' embeddings that modified their activities in response to ToM features, offering initial evidence of a parallel between the artificial model and neurons in the human brain.

## Introduction

In recent years, the rapid evolution of Large Language Models (LLMs) has opened a new era of machine intelligence (*1, 2*). Beyond their remarkable power in language generation, these LLMs have exhibited certain level of competencies across diverse domains, including conversation, code generation, basic mathematical calculation, logical reasoning, and problem-solving tasks (*3-7*). Particularly intriguing is their emerged capacity to engage in Theory of Mind (ToM), a cognitive ability essential for attributing mental states and understanding the perspectives of others (*8, 9*). Notably, recent research has shown LLMs that are capable of archieving ToM skills comparable to those of seven-year-olds (*10*). Although other researchers raise questions about the extent to which large language models can comprehend and simulate theory of mind (*11-13*), it is evident that LLMs have achieved a level of ToM capability that far surpasses the capabilities of earlier, smaller-scale language models (*10*).

Theory of mind is a critical cognitive ability through which humans create intricate mental representations of other agents and comprehend that these agents may possess intentions, beliefs or actions differently from one's own or the objective reality (*8, 9*). A critical test for ToM is the false belief task, which evaluates whether one can recognize that someone may hold an invalid belief that diverges from reality after a change to the environment that they did not witness (*14-16*). For example, a person might believe an apple is still on the tree if that person did not witness the apple falling. Over the past few decades, human brain imaging studies have provided substantial evidence for the brain network that supports our ToM ability, including the temporal-parietal junction, superior temporal sulcus and the dorsal medial prefrontal cortex (dmPFC) (*17-20*). Recently, our research has revealed a detailed single neuronal process in the human dmPFC for representing other's beliefs and identified candidate neurons that could support ToM (*9*). Nevertheless, it remains to be seen whether there exist any parallel for the neural activities associated with human theory of mind in large language models.

Here, we employed a similar methodology employed in human (*9*) to examine the relationship between single neurons in the human brain and the embeddings in the LLM substructures. We aim to begin studying whether and what processes may commonly subserve ToM ability, how they align with task performance, and how they precisely relate to network structure and size. Utilizing open-source LLMs, our initial approach involved a detailed evaluation across multiple ToM tasks, with task materials closely resembling those provided to human participants. Building on these comparisons, we then explored what specific aspects of the hidden embeddings correlated with task performance and the ability of the LLM models to accurately discern false from true beliefs. These results were then compared to those previously obtained from single neurons within the human brain. Finally, we verified our findings by conducting decoding analysis to directly predict the other's beliefs from hidden embeddings. These analyses, in combination, provide insight into how LLMs achieve high-level ToM capabilities, how the hidden network processes involved, and how these compare to those of native biological neurons processing the same precise tasks.

## Results

**Large language models' performances on theory of mind questions**

To first evaluate the capacity of LLMs for ToM, we used four independently trained, open-source LLMs: Falcon (*21, 22*), LLaMa (*23*), Pythia (*24*) and GPT-2 models (*25*). Among them, Falcon and LLaMa exhibited remarkable performance among the open-sourced models, as demonstrated by their rankings on the Huggingface leaderboard (*26*). Each tested LLM encompassed multiple versions with various numbers of hidden layers and parameters, and fine-tuned on multiple datasets, as summarized in **Table 3**. These variations of a model group spanned a broad range of the model performance on language tasks, forming a comprehensive collection of models exhibiting linguistic capabilities.

We initially assessed these models' ability in performing theory of mind tasks using the same time-aligned materials obtained from neuronal recordings as well as how performance was precisely affected by LLM size (**Table 1**) (*9*). Each model underwent independent evaluation through a series of trials comprising a scenario statement followed by two corresponding questions. The statements were designed in pairs with a true belief trial and a false belief trial based on whether the agent's belief matched the reality or not (**Fig. 1A, Table 1**). For example, the statement may provide the scenario "Ned and you take a photo of an apple on a tree. While the photo develops, Ned leaves and is unaware that a wind blows the apple to ground." Since Ned's belief on the location of the apple is different from the reality, this is a false-belief trial. In comparison, a true-belief trial included a statement that Ned's belief is the same as reality (Fig. 1A). The statements were followed by two questions, one relating to the belief of the agent in the scenario statement (i.e., 'belief' question) and the other concerning the physical state of reality (i.e., 'fact' question). To obtain plausible responses from models with different language capabilities, we formulated ToM questions by presenting partial sentences that would guide the predicted word towards being the answer (**Fig. 1A**), and compared the predicted probabilities of the possible words ("tree" or "ground" in this example) to assess whether the correct answer had higher probability than the other (details in **Methods**). Together, our task material is composed of 76 trials. The lengths of the statement varied between 81 words to 191 words, with an average of 125 words.

Overall, we find the tested LLMs had higher accuracies when asked about the facts and others' beliefs in true-belief trials compared to the false-belief trials (**Fig. 1B, C**). Specifically, the accuracies of the predicted answers for the belief questions from the true-belief trials by different LLMs reached an average of 68% (50% chance performance; ranged from 56% to 77%), which was similar to the prediction accuracies on the fact questions (ranged from 55% to 79% with an average of 70%). The false-belief accuracies were lower, by contrast, with an average of only 52% (ranged from 26% to 69%). For these trials particularly, larger models (model parameters ≥12b) performed significantly better than smaller models (≤ 7b, T-test, statistics = 2.88, p = 0.01), with LLaMa-33b model showing the highest accuracy at 69%. In comparison, smaller models showed accuracies lower or similar to chance level. Therefore, although most models exhibited high accuracies to questions about facts or in true-belief trials, only large models showed high accuracies in response to other-belief questions in false-belief trials.

To ensure that the observed accuracies did not independently originate by any clues outside of the scenarios in the statements, we performed the following controls. Firstly, we input each model with the same questions as before, but here we excluded the preceding statements. This control condition therefore allowed us to assess whether factors such as imbalanced word frequencies or linguistic information within the questions could account for the high accuracies. We found that the question-only tests, however, returned an average accuracy of 47% for all models (i.e., chance-level accuracy), with the larger models showing similar performance as the smaller models (T-test, statistics = -0.98, p = 0.34). Secondly, to examine whether the high accuracies may be accounted by factors unrelated to the content of the statement, we randomly permutated words from the statements for each true and false belief trial (**Methods, Table 2**). This resulted in an average accuracy of 55% for all models, and there was no difference between the large and small models for the false belief questions (T-test, statistics = -1.94, p = 0.07). Therefore, these control conditions provided additional confirmation that the remarkable performance of the large models depended on the content of the statements, ruling out explanations based on random factors or word frequency alone.

**Embeddings' selectively tuned to true and false beliefs**

Within human cognition, ToM performance is thought to be supported by a vast network of interconnected neurons that presumably function together to form representations about another's beliefs. Our recent study has identified single neurons in the dorsal medial prefrontal cortex that exhibit selective modulations for true- versus false-belief trials during the period of questions, suggesting a particular role for processing others' beliefs and potentially subserving ToM ability (*9*). Here, we obtained data from single-neuronal recordings from human subjects as they performed a structured false-belief task. Out of 212 recorded human neurons, 49 (23%) displayed significant changes in activities for true- or false-belief trials when human participants performed ToM tasks (**Fig. 2A**). That is, these neurons displayed a consistent difference in their firing rates when the other's beliefs were true compared to when the other's beliefs were false. These neurons therefore reliably changed their activities in relation to the other's beliefs despite variations in the specific statements and scenarios within each trial type, providing evidence for the specific tuning of human neurons to ToM computations.

To investigate whether the artificial models' theory of mind capability shared similar mechanisms as in the human brain, we performed element-wise analysis using the following procedures: Firstly, to obtain the activities of 'artificial neurons' in LLMs, we used hidden embeddings (output of transformer modules) from all layers as well as the input to the first transformer module. Thus, for example, instead of using the firing rate values for each neuron to determine their response selectivity to false versus true beliefs, we used the embedding values for each node in the network (**Methods**). Secondly, to establish a meaningful comparison with human neurons, we employed ToM task materials for LLM closely aligned to the one we tested on humans. Here, we used the same statements as in model evaluation, with trials grouped into pairs of true and false belief trials, and asked a belief question following the statement (**Fig. 2A, Table 1, Method**). These questions were exactly the same for each pair, but the answer depended on the information in the statements which defined the trial types. We modified the statements so that each true-false-belief pair contained similar number of words to minimize any effect caused by variations of word counts. Finally, we input the model with the concatenation of the statement and the question as one batch and only examined the embeddings from the tokens within the questions (detailed explanation in **Method**). We then examined whether embeddings showed significant differences in values between true- and false-belief trials using a Mann Whitney U test. Thus, if an embedding encoded no ToM attributes and solely reflected the literal wording information (which was very similar within each pair) or had no memory of the statements, it would result in similar values between the pair of the trials. Together, the LLM model's hidden embeddings can be thought of, in turn, as the activities of artificial neurons across all network layers that vary in relation to the task and trial-aligned input.

Using this approach, we indeed observed the presence of embeddings with significant responses corresponding to the different trial types. The percentage of the modulated embeddings varied across models and the layers (**Fig. 2B-D**). For example, in the Falcon 40b model, we found 6.3% of significant embeddings in layer 25, which represented the highest percentage among the layers. These embeddings showed either increased or decreased activities for true- versus false-belief trials (**Fig. 2B**). By contrast, there was no responsive embedding from the input layer up to the layer 8 in this model (**Fig. 2D** *left, right inset*). A similar pattern was observed in the LLaMa-30b models (**Fig. 2D** *left, middle inset*), in which 5.6% of embeddings at 19$^{th}$ layer exhibited selectivity to trial types, and very few were responsive from the input up to the 9$^{th}$ layer. This trend of significant artificial neurons present in the middle and high layers was consistent across models.

Next, we assessed the percentage of embeddings displaying significant selectivity from various models by using the percentage from the layer with the highest percentage of each model. In general, the percentage of significant embeddings increased with the model size (**Fig. 2D** *left*). For large models ($\geq$ 12b), there was an average of 3.9% of embeddings responding to ToM tasks, and this percentage dropped to 0.6% for smaller models (T test, statistics = -4.6, p = 4 x 10$^{-4}$). Collectively, the percentage of significant embeddings were also closely correlated to the model performance (**Fig. 2D** *right*). For models with above-chance performance, the percentage of ToM-responsive embeddings increased non-linearly, with an exponential relation between the percentage and the performance (percentage = $a$ exp ($b \cdot$ performance), where $a = 0.01 \pm 2.1$ x 10$^{-5}$, $b = 6.1 \pm 4.4$). Together, our findings revealed the presence of embeddings that displayed modulations related to the theory of mind content in multiple large models, a feature that was absent in smaller models with chance-level false-belief performance.

Finally, to ensure the above findings cannot be explained by random fluctuation or other features unrelated to the ToM information in the statements, we conducted a control experiment by randomly permuting words in the statements. We then applied the same criterion to select responding embeddings. We found that the percentages were significantly lower compared to those resulted from the intact statements for large models (T-test, statistic = 4.1, p = 0.002) but not for small models (T-test, statistic = 1.46, p = 0.16). These, together, indicated that the presence of ToM responsive neurons in the large models cannot be explained by clues unrelated to the contextual information in the scenario statements. Therefore, although the percentage of ToM artificial neurons were considerably lower than those observed in the human brain (23%), there was an emergence of 'artificial' neurons in middle and high layers of the large LLMs that responded to ToM features.

**True and false beliefs can be decoded from the entire embeddings**

Next, to further investigate the relationships between the hidden embeddings and the models' ToM capability, we examined whether others' beliefs (i.e., true vs false beliefs) can be directly decoded from the population of hidden embeddings. Specifically, we used all dimensions of embeddings derived from each layer within a given model, and trained a logistic regression with L2 regularization to predict the trial types for trials that were not in the training dataset (details in **Methods**). Here, we find a majority of the true- and false-belief trial types were accurately decoded using the entire hidden embeddings from the 25$^{th}$ layer of the Falcon 40b model (**Fig. 3A** *top*). Furthermore, the activities of significant neurons exhibited far greater discrimination between false and true belief trials in correctly decoded trials compared to incorrectly decoded trials (average z-scored differences were 0.60 and 0.25, respectively; T-test, statistic = 17.9, p =

1.6 x 10$^{-62}$, **Fig. 3A** *bottom*). Together, the activities of these artificial neurons therefore appeared to be predictive of the model's ToM performance.

Examining all models together, the decoding accuracies increased with the size of the models, with large models (≥ 12b) showing an average of 75% decoding accuracy. The Falcon-40b model showed the highest decoding accuracy of 81%. The embeddings in smaller models (≤ 7b), however, could only predict the trial types at an average accuracy of 67%, which was significantly lower than those from the large models (T-test, statistic = -4.2, p = 0.001). This observation was also consistent with the ratio of responding neurons, together suggesting a relation between the size of the models and the proportion of artificial neurons capable of accurately predicting the other's beliefs.

Finally, to ensure that the decoding accuracies were not originated from factors unrelated to the scenario statements, we randomly permuted the words in each pair of the statements and repeated the same decoding procedures to decode the trial type (**Methods**). Here, the decoding accuracies from all models dropped to an average of only 55%, which was significantly lower than all accuracies without the random permutation (T-test, $p < 3 \times 10^{-110}$). The differences of accuracies between the intact and permuted control were higher for large models, with an average of 19%. These findings showed that the ToM trial types can be robustly decoded from the population of artificial neurons (embeddings), indicating a consistent encoding of ToM features by the embeddings. Together with the results from individual embedding, our results collectively support the hypothesis that hidden embeddings possess the capacity to effectively predict the other's beliefs, suggesting their role in facilitating the models' ToM performance.

## Discussion

The ability to discern between true and false beliefs represents a significant aspect of theory of mind that is proposed to be linked to our conscious mind (*27, 28*). Recent advancements in large language models (LLMs) have revealed their potentials in distinguishing objective reality from false beliefs (*10, 12*). Our study aims to provide an initial investigation to the possible mechanisms underlying ToM in LLMs. By analyzing hidden embeddings from various open-source LLMs, we uncovered the presence of hidden embeddings that were predictive of the beliefs of others across richly varied scenarios. This finding is particularly remarkable, considering that the embeddings were derived from identical questions following narratives with very similar wording. This suggests the models' ability to not only differentiate subtle variations among closely related sentences, but also categorize them based on true and false beliefs, thereby encoding the perspective of others. These responses were absent when we randomly permuted the words in statements while keeping the questions intact. Additionally, the trial types (i.e., true- or false-belief) were accurately decoded from the population of embeddings, further validating the robust representation of ToM within the artificial models. Finally, we observed a strong and positive relation between the task performance and the proportion of ToM-responsive embeddings, suggesting their role in facilitating the performance. Collectively, our findings indicate an emergence of ToM-related embeddings in the artificial models, supporting the model capability in capturing essential aspects of ToM.

Although, unlike humans, LLMs were trained solely on language materials and lacked rich resources by which humans develop ToM capability (*29, 30*), the emergent behavior observed in the artificial models bears a striking resemblance to the neuronal activity associated with ToM in the human brain. With hidden embeddings as counterparts of brain neurons, both systems contain neurons that directly respond to the perspective of others. We showed that a substantial proportion of artificial neurons that responded selectively to true- or false-belief trials, mirroring prefrontal neurons in humans exhibiting changes in firing rates for different trial types (*9*). Furthermore, the LLM layers with high percentages of ToM-responding embeddings were consistently not confined to one or two layers or distributed randomly. Rather, they showed a peak in the middle and high layers and almost zero in the input layers. A similar distributed areas for ToM were observed in human brain, particularly within areas of the frontal, temporal and parietal cortices (*9, 17-20*), which have been identified as regions for high-level cognitive processing. ToM-related activity within lower input processing areas such as occipital lobe is minimal. Finally, we observed the artificial layers exhibiting ToM responses were located in contiguous layers, analogous to the highly interconnected structure of ToM brain areas. Altogether, these observations are remarkable because humans rely on many years of development and real-world social interactions with others to form ToM capability (*29, 30*). The LLMs tested here, by comparison, are largely trained on vast language corpora with no explicit experience in interacting with others or direct representation of agency. Yet, despite significant structural and algorithmic differences between the artificial and brain networks, they indeed exhibit surprising convergence by adopting similar mechanism of encoding ToM information. This convergence is evident both in their capability to differentiate true and false beliefs and in the emergence of ToM-related neurons that facilitate such cognitive functions.

Collectively, these results shed light on the potential of large language models to exhibit theory of mind capabilities and contribute to our understanding of cognitive processes in artificial intelligence. However, our findings are limited to open-source LLMs, as we did not have access to the hidden embeddings of the higher-performing LLMs such as GPT-4 (*7*), which could offer further insights into the relationship between model performance and embedding representation. Further, our methods excluded embeddings that were selective to both true- and false-belief trials and only focused on the embeddings that showed selectivity to one of them. Nevertheless, our findings represent the initial exploration into the role of embeddings in ToM within language models and provide insights in how artificial intelligence can exhibit sophisticated cognitive abilities.

## Methods

### Theory of mind (ToM) materials

To assess the artificial language models' capacity for theory of mind and to ensure a direct comparison with human performance, we used testing materials previously employed in human studies during single neural recordings. Minor adjustments were made to accommodate the specificities of artificial models (e.g., statements in pairs were slightly modified to have similar lengths). The ToM ability of each model was evaluated using 76 trials consisting of a scenario statement followed by two related questions: a 'belief question' related to the belief of the agent in the scenario statement and a 'fact question' concerning the physical state of reality (**Fig. 1, Table 1**). Across all trials we presented, the lengths of the statement varied between 81 words to 191 words, with an average of 125 words.

*Scenario statements.* The trials were grouped in pairs, containing one true-belief and one false-belief trial in each pair. The trials in a pair start with very similar scenario statements providing background for the reader to infer whether the agent's belief in the story is aligned with the reality or not (true-belief or false belief, respectively; see examples in Table 1). In addition, we ensured each pair of true- and false-belief trials contain the same number of words in the statements, so that the potential variances stemming from different word positions in the sentence are minimized.

*Questions for evaluating model performance.* Based on the statements described above, we designed two categories of questions to test the ToM capability of the large language models (LLMs): a fact question and an other-belief question (**Table 1**). We edited the structure of the question in order to obtain an objective evaluation of the model ability. For example, after a scenario statement like "Charles left his wallet on the counter as he was leaving the store. The wallet fell on the floor. Charles returns", if we asked 'Where will Charles look for the wallet?', an LLM might generate a long paragraph without directly answering the question, making it subjective to assess whether the model answered the question correctly or not. Here, given that all LLMs we assessed generate outputs in the form of predicted upcoming words with a probability distribution across all possible tokens, we modified the questions to align with this characteristic of the LLMs. In the example provided above, we asked "Charles will look for the wallet on the". In this way, LLM models will likely predict a location for the upcoming word.

*Question for evaluating others' belief processing by hidden embeddings.* Here, the goal of these questions is not to evaluate model performance, but to examine whether hidden embeddings show selectivity to the trial types (false-belief or true-belief), and to directly compare the results to those from single neurons in human brains. Therefore, we used the same set of questions as those posed to human participants to ensure reasonable comparison with findings from single neurons recorded in prefrontal cortex of human brains. Specifically, we asked the same belief questions for each pair of true- and false-belief trials, using the same format in (*9*), e.g., "Where will Charles look for his wallet?" In this way, the pair of true- and false-belief trials were composed with very similar words and with exactly the same questions (**Table 1, Fig. 2**).

Table 1. Example of the task materials

| Trial type | Statement | Fact question | Belief question | Belief question in the human study |
|---|---|---|---|---|
| False belief | Mary put fish inside a jewelry box while her son wasn't looking. Her son opens the box. | Inside the box, there is | Inside the box, he expects to find | What does he expect to find? |
| True belief | Mary put jewelry inside a jewelry box and her son sees it. Her son opens the box. | Inside the box, there is | Inside the box, he expects to find | What does he expect to find? |

| Trial type | Statement | Fact question | Belief question | Belief question in the human study |
|---|---|---|---|---|
| False belief | Ned and you take a photo of an apple on a tree. While the photo develops, Ned leaves and is unaware that a wind blows the apple to ground. | Currently, the apple is on the | Ned believes that the apple is on the | Where does Ned believe the apple is? |
| True belief | Ned and you take a photo of an apple on a tree. While the photo develops, you and Ned see a strong wind blow the apple on the ground. | Currently, the apple is on the | Ned believes that the apple is on the | Where does Ned believe the apple is? |
| False belief | Charles left his wallet on the counter as he was leaving the store. The wallet fell on the floor. Charles returns | The wallet is on the | Charles will look for the wallet on the | Where will Charles look for the wallet? |
| True belief | Charles left his wallet on the counter as he was leaving the store. No one has touched his wallet. Charles returns. | The wallet is on the | Charles will look for the wallet on the | Where will Charles look for the wallet? |

**Control tasks**

To ensure our observations are not derived from factors unrelated to the scenario created in the statements, we performed the following two controls. First, we created shuffled control trials by randomly permutating words in each statement while keeping the questions intact (**Table 2**). In this way, we kept the same words in the statement but eliminated the contextual information. Second, we estimated the impact of any clues within the questions (e.g., the potential imbalance of word frequency) by inputting each model with the questions only. The combination of these two controls will provide estimation on the impact of factors unrelated to the ToM-related content provided by the statement.

Table 2. Example of control task by random shuffle words in statement

| Trial type | Statement | Fact question | Belief question | Belief question in the human study |
|---|---|---|---|---|
| False belief | her son jewelry Mary looking. Her fish son put while box inside wasn't opens the box. a | Inside the box, there is | Inside the box, he expects to find | What does he expect to find? |
| True belief | inside Her and the box it. Mary her box. jewelry a opens son put jewelry sees son | Inside the box, there is | Inside the box, he expects to find | What does he expect to find? |
| False belief | and take the photo the a wind an Ned Ned leaves tree. apple on is unaware a photo blows and develops, ground. While of you apple a to that | Currently, the apple is on the | Ned believes that the apple is on the | Where does Ned believe the apple is? |
| True belief | While on you develops, the on you the Ned apple blow the apple an tree. Ned and take and photo a ground. strong a wind of see a photo | Currently, the apple is on the | Ned believes that the apple is on the | Where does Ned believe the apple is? |
| False belief | on store. his left as the counter leaving was The wallet returns on the Charles wallet floor. fell Charles the he | The wallet is on the | Charles will look for the wallet on the | Where will Charles look for the wallet? |

| True belief | has No his one counter store. the returns. as on wallet wallet. Charles Charles the was he leaving touched his left | The wallet is on the | Charles will look for the wallet on the | Where will Charles look for the wallet? |

## Large language models (LLMs)

Our study primarily focuses on four high-performing, independently trained language models that are publicly available as open source. All LLM models examined were composed of transformer modules that were connected in sequence. Each LLM contains multiple versions, characterized by varying numbers of parameters and potential fine-tunning on specific datasets and training. Specifically, these models include Falcon (1b, 7b, 40b), llama (3b, 7b, 13b, 30b, 33b), Pythia (3b, 7b, 12b), and GPT-2 (medium, large, xl). The tokenizers and parameters from all models were downloaded in July 2023 and were not updated since then. The details of the model information and the datasets that they were fine-tuned on are listed in **Table 3**. In our study, all models and tokenizers were loaded via Huggingface in Python (*31*). For models with a parameter count of less than or equal to 7b, we utilize a desktop computer with single GPU (NVIDIA GeForce RTX 4090). For larger models, we utilize the Massachusetts General Hospital GPU cluster facility with up to eight GPUs (NVIDIA DGX-1) for model performance and evaluations.

Table 3. Large language models examined in this study

| Model name | Base model | Model source | Size | Description from model developer |
|---|---|---|---|---|
| Falcon-1b | Falcon | tiiuae/falcon-rw-1b | 1b | Decoder model; Trained on 350B tokens of RefinedWeb (*22*) |
| Falcon-7b | Falcon | tiiuae/falcon-7b | 7b | Decoder model; Trained on 1,500B tokens of RefinedWeb; Enhanced with curated corpora. |
| Falcon-40b | Falcon | tiiuae/falcon-40b-instruct | 40b | Decoder model; Based on Falcon-40B; Finetuned on a mixture of Baize. |
| LLaMa-3b-1 | LLaMa | openlm-research/open_llama_3b | 3b | An Open Reproduction of LLaMA (*32*) |
| LLaMa-7b-1 | LLaMa | openlm-research/open_llama_7b | 7b | An Open Reproduction of LLaMA |
| LLaMa-13b-1 | LLaMa | ausboss/llama-13b-supercot | 13b | Merge of LLAMA-13b and SuperCOT LoRA (*33*) |
| LLaMa-30b-1 | LLaMa | ausboss/llama-30b-supercot | 30b | Supercot; Work with langchain prompting |
| LLaMa-7b-2 | LLaMa | eachadea/vicuna-7b-1.1 | 7b | Chatbot; Fine-tuned on user-shared conversations from ShareGPT (*34*) |
| LLaMa-13b-3 | LLaMa | openaccess-ai-collective/wizard-mega-13b | 13b | Fine-tuned on the ShareGPT, WizardLM, and Wizard-Vicuna datasets (*35*) |
| LLaMa-33b-4 | LLaMa | elinas/chronos-33b | 33b | Focused on chat, roleplay, and story-writing (*36*) |
| Pythia-3b | Pythia | databricks/dolly-v2-3b | 3b | Trained on the Databricks machine learning platform (*37*) |
| Pythia-7b | Pythia | databricks/dolly-v2-7b | 7b | Trained on the Databricks machine learning platform |

| Pythia-12b | Pythia | databricks/dolly-v2-12b | 12b | Trained on the Databricks machine learning platform |

## Evaluating ToM performance

Using the ToM materials described above, for each trial, we concatenated the statement and the corresponding question and fed them into the model. From the model output, we obtained the model's prediction of the next word by examining the output logits of all possible tokens. These logits are monotonically related to the probability of the upcoming word predicted by the model. Then we specifically examined the logits of the two possible answers for the belief and fact questions. To determine the LLMs' answer, we chose the word with the highest logit value out of the two word choices (e.g. floor or counter) to ensure the selection of more reliable predictions and avoid instances where certain models generate irrelevant outputs. Then the accuracies of each model were calculated for true beliefs, false beliefs, and facts by considering all trials with corresponding questions. The same procedures were followed for the two control conditions described above to further verify our findings.

## Hidden embeddings' selectivity for true- or false- belief trials

For each LLM, we tested their ToM capacity by extracting the model's hidden embeddings from all layers along with the predicted logit for the upcoming token during the task. Specifically, for each trial, we concatenated the statements and the questions we presented to the human participants as a single input to the LLM model (**Table 1, Fig. 2a**). The hidden embeddings were obtained from the output of each transformer module, in addition to the one that was input to the first transformer module for each model. The dimension of the returned embeddings for each model in this step was trials x words x nodes x layers, where words included those from both the statement and the question, and nodes referred to the embedding size of a layer. Following a comparable approach employed to evaluate ToM-related activities from single neurons in the human brain, we used the embeddings corresponding to the question tokens and subsequently calculated the average values across all question tokens (dimension of trials x nodes x layers). We then performed statistical tests to evaluate whether each embedding (looping over nodes and layers) exhibited significant responses to trial conditions (i.e., true-belief and false-belief). Particularly, we compared the embedding values between these two trial conditions with the Mann Whitney U test, testing the null hypothesis that the distributions of the two categories were the same. The statistic for the Mann Whitney U Test is the minimum of $U_1$ and $U_2$, defined as

$$U_1 = n_1 n_2 + \frac{n_1(n_1 + 1)}{2} - R_1$$
$$U_2 = n_1 n_2 + \frac{n_2(n_2 + 1)}{2} - R_2$$

where $R_1$ and $R_2$ are the sum of the ranks for group 1 and 2 respectively. We used a threshold of 0.05 to determine whether a given dimension of an embedding demonstrated significant association to the trial category. Next, we grouped the embeddings based on their layers and models, and calculated the percentage of embeddings that showed higher than chance responsiveness. We examined all embeddings across different layers within a given LLM, then for each model, we selected the layer with the highest percentage of responsive embeddings as the percentage of that model. All steps described here were repeated for the control experiments with the random permuted words in the statements described above for further verification.

**Decoding the trial type using the population of embeddings**

In order to examine whether there is a causal relationship between the observed selectivity of the embeddings and model performance, we conducted decoding analysis using the entire population of embeddings of each layer. Specifically, for each layer, from the embeddings with the dimension of trials x words x nodes, we averaged across question tokens for each trial for a given layer of each model, resulting in the dimension of trials x nodes. We considered nodes as the equivalent of neurons in the brain to predict the type of trials as the target variable. We used 75% training and 25% testing split based on the pair of trials, so that trials within a pair were not separated into two datasets. We used a logistic regression classifier with L2 regularization of $C = 1$, which minimize the cross-entropy loss with the penalty of the square of the parameter values:

$$\min C \left( \sum_{i=1}^{n} \left( -y_i \log(\hat{p}(X_i)) - (1 - y_i) \log(1 - \hat{p}(X_i)) \right) \right) + \frac{1}{2} \|w\|^2$$

where the target variable $y_i$ belongs to the set $\{0, 1\}$ for data point $i$, and $w$ is the weight.

For each layer of a given LLM, we performed the same analysis 100 times with different train and test splits, and calculated the average accuracies across these iterations. At the end, the decoding accuracy of each model was calculated by taking the average over all layers from the model. As a control, we repeated the same procedures for the same layer, but using the ToM materials with the randomly permuted words in the statements.

## Acknowledgement

We are grateful to Yuedong Fang, Yue Cao and Nikola Bolt for their comments to improve the manuscript, and Douglas Kellar for facilitating access to computational resources. We acknowledge the utilization of ChatGPT for assistance in refining the wording of this manuscript (*38*). Z.M.W. is supported by NIH U01NS123130.

## Code availability

All codes will be made publicly available on GitHub when the manuscript is accepted for publication.

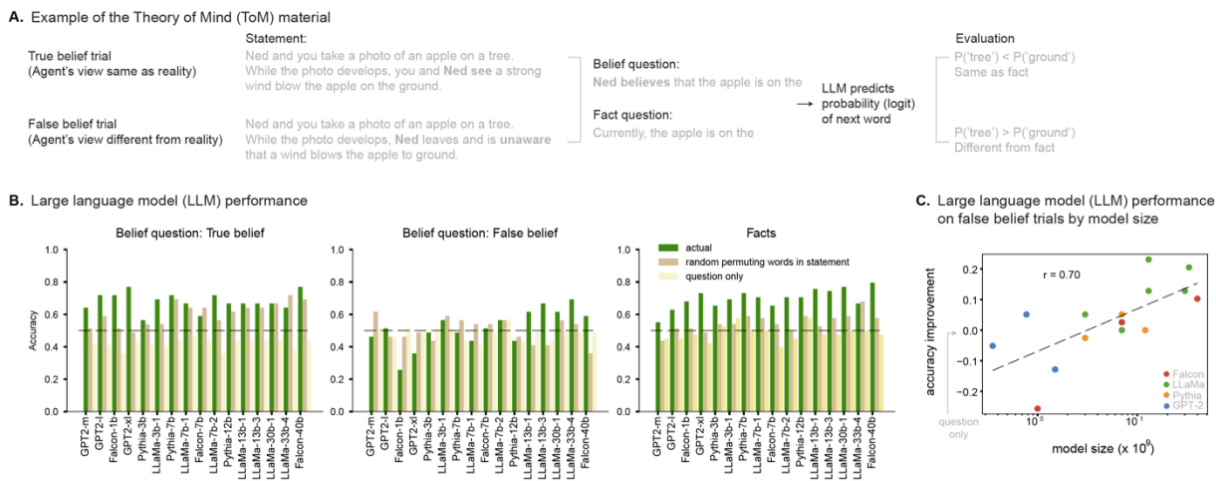

**Figure 1. Theory of Mind capability in various large language models. A.** ToM tasks comprising statements and questions were input to each LLM, and the predicted upcoming word and probabilities were examined (**Method**). The ToM trials were either true- or false-belief, depending on whether the agent's viewpoint was aligned with the reality or not. Besides, we assessed the models' ability to answer questions about the factual state of reality provided by the statements. **B.** Model performance on questions of true-belief (*left*), false-belief (*middle*), and facts trials (*right*). For control experiments, we randomly permutated words within the statements and input these shuffled words along with questions to the models and repeated the same model evaluation procedures. We also assessed models' performance on questions-only trials without the statements to evaluate impact of factors unrelated to the context provided by the statements. **C.** LLMs' accuracies in answering false-belief questions and their dependency on the size of the models. We plotted the accuracy improvement resulting from inputting statements and questions compared to accuracy from only inputting questions, across different LLMs.

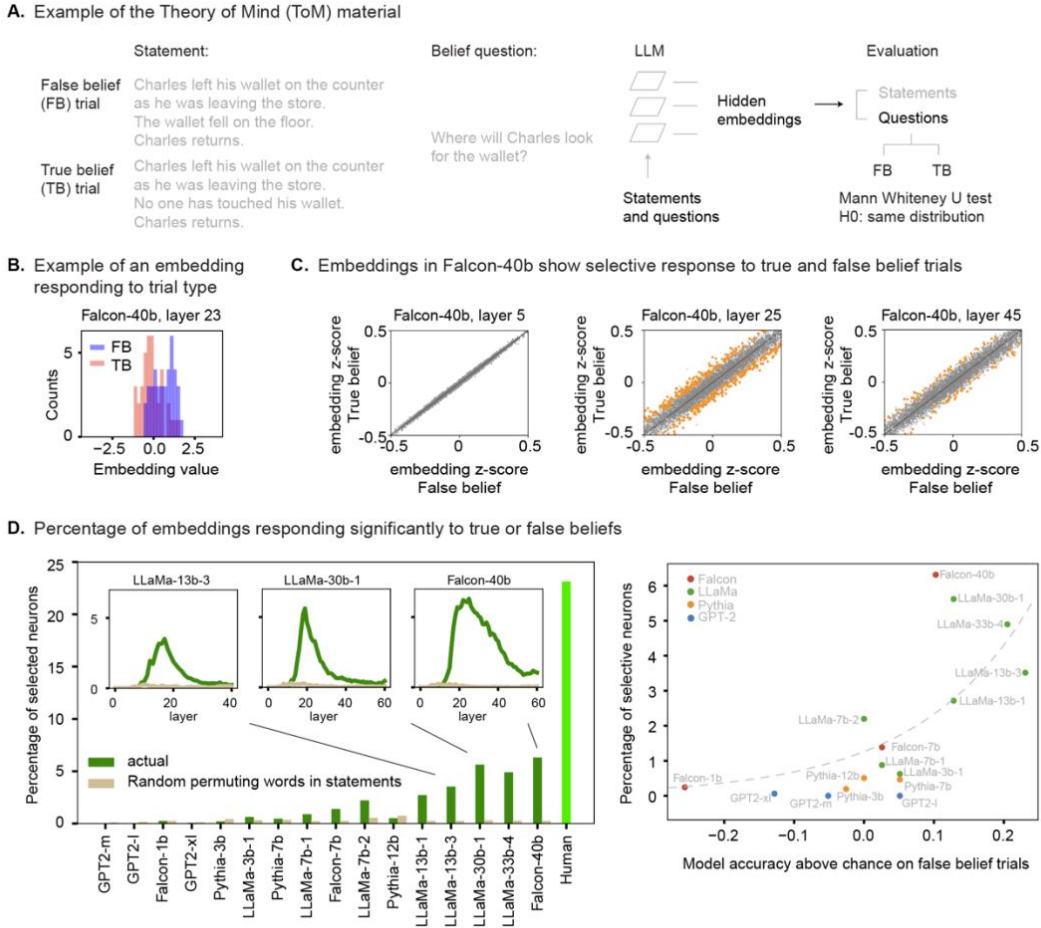

**Figure 2. Responding embeddings to true- versus false-belief trials. A.** To investigate whether hidden embeddings exhibit selective modulations to true versus false beliefs and to compare the artificial embeddings with human single neurons, we employed similar ToM tasks as those previously tested on humans. For each trial, the statement and the question were concatenated and input to the LLMs, obtaining hidden embeddings from all words and layers (**Methods**). We then excluded embeddings from words during the statement and computed average values for words within the question. A Mann Whitney U test was conducted to examine whether a hidden embedding exhibited significant difference between false- and true-belief trials. **B.** Distributions of embedding values from Falcon-40b layer 23 to illustrate that the activities were significantly different for false-belief trials than for true-belief trials. **C.** Examples from different layers of the Falcon-40b model show the average embedding values over true- and false-belief trials. Each dot represents a dimension of the hidden embedding; orange dots indicate the embedding with significant differences between trial types, while the gray dots indicate no significance. **D.** The percentage of embedding dimensions significantly selective to true and false beliefs varies across models and layers (*left*), with the Falcon-40b model demonstrating the highest percentage. These results are compared to the percentage of single neurons in the human brain (light green). The percentages across layers of three example models are shown in the *insets*. The percentages of significant embeddings across different models were found to be dependent on the false-belief trial accuracy (*right*).

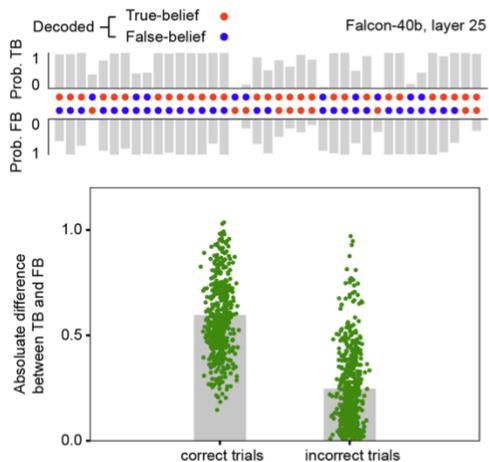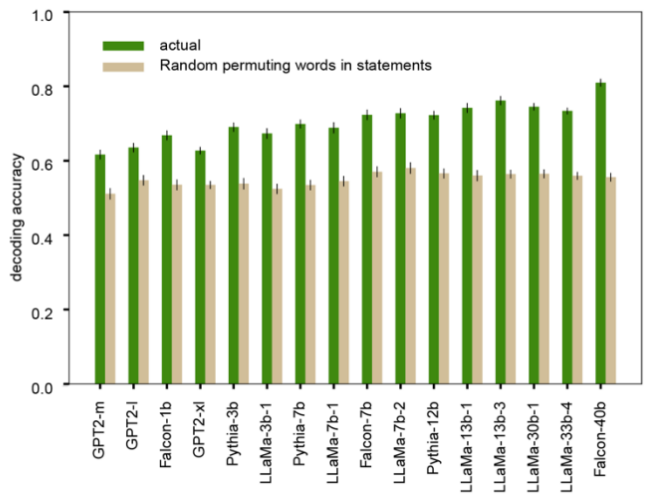

**Figure 3. Decoding trial types using hidden embeddings. A.** Using Falcon-40b as an example, higher probabilities of the correct trial type were observed from most observations decoded from all embeddings at layer 25 (*top*). The selected embeddings showed greater difference between true- and false-belief trials in correctly decoded trials compared to the incorrectly decoded trials (*bottom*). **B.** Across different models, large models generally demonstrated higher decoding accuracy for true- and false-belief trials using all embeddings from each layer. In contrast, decoding accuracies remained consistently low when the words in the statements were randomly permuted before inputting them into the LLMs.

# Reference:

1. N. Aggarwal, G. J. Saxena, S. Singh, A. Pundir, Can I say, now machines can think? *arXiv preprint arXiv:2307.07526*, (2023).
2. M. Sallam, in *Healthcare*. (MDPI, 2023), vol. 11, pp. 887.
3. J. He-Yueya, G. Poesia, R. E. Wang, N. D. Goodman, Solving math word problems by combining language models with symbolic solvers. *arXiv preprint arXiv:2304.09102*, (2023).
4. Z. Yuan, H. Yuan, C. Tan, W. Wang, S. Huang, How well do Large Language Models perform in Arithmetic tasks? *arXiv preprint arXiv:2304.02015*, (2023).
5. L. Pan, A. Albalak, X. Wang, W. Y. Wang, Logic-lm: Empowering large language models with symbolic solvers for faithful logical reasoning. *arXiv preprint arXiv:2305.12295*, (2023).
6. S. Yao *et al.*, Tree of thoughts: Deliberate problem solving with large language models. *arXiv preprint arXiv:2305.10601*, (2023).
7. OpenAI, GPT-4 Technical Report. *ArXiv* **abs/2303.08774**, (2023).
8. C. Frith, U. Frith, Theory of mind. *Current biology* **15**, R644-R645 (2005).
9. M. Jamali *et al.*, Single-neuronal predictions of others' beliefs in humans. *Nature* **591**, 610-614 (2021).
10. M. Kosinski, Theory of mind may have spontaneously emerged in large language models. *arXiv preprint arXiv:2302.02083*, (2023).
11. T. Ullman, Large language models fail on trivial alterations to theory-of-mind tasks. *arXiv preprint arXiv:2302.08399*, (2023).
12. S. Trott, C. Jones, T. Chang, J. Michaelov, B. Bergen, Do Large Language Models know what humans know? *Cognitive Science* **47**, e13309 (2023).
13. M. C. Frank, Baby steps in evaluating the capacities of large language models. *Nature Reviews Psychology*, 1-2 (2023).
14. H. M. Wellman, D. Cross, J. Watson, Meta-analysis of theory-of-mind development: The truth about false belief. *Child development* **72**, 655-684 (2001).
15. H. Wimmer, J. Perner, Beliefs about beliefs: Representation and constraining function of wrong beliefs in young children's understanding of deception. *Cognition* **13**, 103-128 (1983).
16. K. Milligan, J. W. Astington, L. A. Dack, Language and theory of mind: Meta-analysis of the relation between language ability and false-belief understanding. *Child development* **78**, 622-646 (2007).
17. V. E. Stone, S. Baron-Cohen, R. T. Knight, Frontal lobe contributions to theory of mind. *Journal of cognitive neuroscience* **10**, 640-656 (1998).
18. M. Siegal, R. Varley, Neural systems involved in 'theory of mind'. *Nature Reviews Neuroscience* **3**, 463-471 (2002).
19. R. Saxe, N. Kanwisher, People thinking about thinking people: the role of the temporo-parietal junction in "theory of mind". *Neuroimage* **19**, 1835-1842 (2003).
20. R. Saxe, L. J. Powell, It's the thought that counts: specific brain regions for one component of theory of mind. *Psychological science* **17**, 692-699 (2006).
21. G. Penedo *et al.*, The RefinedWeb dataset for Falcon LLM: outperforming curated corpora with web data, and web data only. *arXiv preprint arXiv:2306.01116*, (2023).
22. E. Almazrouei *et al.* (2023).

23. H. Touvron *et al.*, LLaMA: open and efficient foundation language models, 2023. *URL https://arxiv. org/abs/2302.13971*.
24. S. Biderman *et al.*, in *International Conference on Machine Learning*. (PMLR, 2023), pp. 2397-2430.
25. A. Radford *et al.*, Language models are unsupervised multitask learners. *OpenAI blog* **1**, 9 (2019).
26. E. Beeching *et al.*, Open LLM Leaderboard. *Hugging Face*, (2023).
27. U. Frith, F. Happé, Theory of mind and self-consciousness: What is it like to be autistic? *Mind & language* **14**, 82-89 (1999).
28. J. Perner, Z. Dienes, Developmental aspects of consciousness: How much theory of mind do you need to be consciously aware? *Consciousness and cognition* **12**, 63-82 (2003).
29. J. I. Carpendale, C. Lewis, Constructing an understanding of mind: The development of children's social understanding within social interaction. *Behavioral and brain sciences* **27**, 79-96 (2004).
30. C. Lewis, N. H. Freeman, C. Kyriakidou, K. Maridaki-Kassotaki, D. M. Berridge, Social influences on false belief access: Specific sibling influences or general apprenticeship? *Child development* **67**, 2930-2947 (1996).
31. T. Wolf *et al.*, Huggingface's transformers: State-of-the-art natural language processing. *arXiv preprint arXiv:1910.03771*, (2019).
32. X. a. L. Geng, Hao. (2023).
33. https://huggingface.co/ausboss/llama-13b-supercot.
34. W.-L. Chiang *et al.*, Vicuna: An open-source chatbot impressing gpt-4 with 90%* chatgpt quality. *See https://vicuna. lmsys. org (accessed 14 April 2023)*, (2023).
35. https://huggingface.co/openaccess-ai-collective/wizard-mega-13b.
36. https://huggingface.co/elinas/chronos-33b.
37. M. Conover *et al.*, Free dolly: Introducing the world's first truly open instruction-tuned llm. (2023).
38. ChatGPT, https://chat.openai.com/chat.